\documentclass[journal]{IEEEtran}

\def\endthebibliography{%
  \def\@noitemerr{\@latex@warning{Empty `thebibliography' environment}}%
  \endlist
}

\usepackage{cite}
\usepackage{svg}
\usepackage{times}
\usepackage{booktabs}
\usepackage{siunitx}
\usepackage{amsmath}
\usepackage{amssymb}
\usepackage{bm}

\usepackage[colorlinks=true, linkcolor=blue, citecolor=blue, urlcolor=blue]{hyperref}
\usepackage{multicol}
\usepackage[nolist,nohyperlinks]{acronym}
\begin{acronym}
    \acro{DoFs}{Degrees of Freedom}
    \acro{CSR}{Continuum Soft Robot}
    \acro{CSM}{Continuum Soft Manipulator}
    \acro{CRT}{Cosserat Rod Theory}
    \acro{ODE}{Ordinary Differential Equation}
    \acro{ISP}{Implicit Strain Parameterization}
    \acro{DER}{Discrete Elastic Rod}
    \acro{GVS}{Geometrically Variable Strain}
    \acro{PCS}{Piecewise Constant Strain}
    \acro{PLS}{Piecewise Linear Strain}
    \acro{PAS}{Piecewise Affine Strain}
    \acro{PCC}{Piecewise Constant Curvature}
    \acro{PAC}{Piecewise Affine Curvature}    
    \acro{SoRoSim}{Soft Robot Simulator}
    \acro{OCP}{Optimal Control Problem}
    \acro{TO}{Trajectory Optimization}
    \acro{DC}{Direct Collocation}
    \acro{NLP}{Nonlinear Programming}
    \acro{IP}{Interior-Point}
    \acro{SQP}{Sequential Quadratic Programming}
    \acro{DP}{Dynamic Programming}
    \acro{DDP}{Differential Dynamic Programming}
    \acro{IDDP}{Implicit Differential Dynamic Programming}
    \acro{Box-DDP}{Box-Differential Dynamic Programming}
    \acro{Box-IDDP}{Box-Implicit Differential Dynamic Programming}
    \acro{Box-FDDP}{Box-Feasibility-driven Differential Dynamic Programming}
    \acro{MPC}{Model Predictive Control}
    \acro{NMPC}{Nonlinear Model Predictive Control}
    \acro{SMC}{Sliding Mode Control}
    \acro{NN}{Neural Network}
    \acro{FEM}{Finite Element Method}

    \acro{CC}{Constant Curvature}
    \acro{VC}{Variable Curvature}
    \acro{CS}{Constant Strain}
    \acro{VS}{Variable Strain}
\end{acronym}

\renewcommand{\baselinestretch}{0.965}

\usepackage{algorithm}
\usepackage{algpseudocode}

\newcommand{\sssabiorob}
{
    The BioRobotics Institute, 
    Scuola Superiore Sant'Anna, 
    Pisa, Italy.
}
\newcommand{\sssarob}
{
    Department of Excellence in Robotics and AI, Scuola Superiore Sant'Anna, Pisa, Italy.
}
\newcommand{\tudelft}
{
    Department of Cognitive Robotics, 
    Delft University of Technology, 
    2628 CN Delft, The Netherlands.
}

\newcommand{\kumech}
{
    Department of Mechanical and Nuclear Engineering, Khalifa University of Science and Technology, Abu Dhabi, UAE.
}
\newcommand{\kurob}
{
    Khalifa University Center for Autonomous Robotic Systems (KUCARS), Abu Dhabi, UAE.
}
\newcommand{\dlr}
{
Institute of Robotics and Mechatronics, German Aerospace Center (DLR), 82234 Oberpfaffenhofen, Germany.
}

\begin{document}
\bstctlcite{IEEEexample:BSTcontrol}

\title{Soft Swing-up: Benchmarking Model-Based Optimal Control for Rigid-Soft Underactuated Systems}

\author
{
    Daniele~Caradonna$^{1, 2}$, \and
    Nikhil~Nair$^{3}$, \and
    Anup~Teejo~Mathew$^{4, 5}$, \and
    Daniel~Feliu-Talegon$^{3}$, \and
    Imran~Afgan$^{4, 5}$, \and
    Egidio~Falotico$^{1, 2}$, \and
    Cosimo~Della~Santina$^{3, 6}$, \and
    Federico~Renda$^{4, 5}$
    \thanks
    {
        $^{1}$\sssabiorob
        $^{2}$\sssarob
        $^{3}$\tudelft
        $^{4}$\kumech
        $^{5}$\kurob
        $^{6}$\dlr
    }
}

\maketitle
\begin{abstract}
Continuum soft robots are inherently underactuated and subject to intrinsic input constraints, making dynamic control particularly challenging, especially in hybrid rigid–soft robots.
While most existing methods focus on quasi-static behaviors, dynamic tasks such as swing-up require accurate exploitation of continuum dynamics.
This has led to studies on simple low-order template systems that often fail to capture the complexity of real continuum deformations.
Model-based optimal control offers a systematic solution; however, its application to rigid–soft robots is often limited by the computational cost and inaccuracy of numerical differentiation for high-dimensional models.
Building on recent advances in the Geometric Variable Strain model that enable analytical derivatives, this work investigates three optimal control strategies for underactuated soft systems—Direct Collocation, Differential Dynamic Programming, and Nonlinear Model Predictive Control—to perform dynamic swing-up tasks. To address stiff continuum dynamics and constrained actuation, implicit integration schemes and warm-start strategies are employed to improve numerical robustness and computational efficiency. The methods are evaluated in simulation on three Rigid-Soft and high-order soft benchmark systems—the Soft Cart-Pole, the Soft Pendubot, and the Soft Furuta Pendulum—highlighting their performance and computational trade-offs.
\end{abstract}


\section{Introduction} \label{sec:introduction}
\acp{CSR} exhibit continuous, highly deformable behavior due to their theoretically infinite \ac{DoFs}, enabling dexterity and adaptability beyond rigid robots. 
Despite their infinite-dimensional configuration space, \acp{CSR} are actuated by a finite number of inputs, rendering them inherently underactuated and significantly complicating modeling and control~\cite {della2023model, falotico2024learning}.

Consequently, most control approaches for \acp{CSR} are limited to quasi-static behaviors, leaving their dynamic potential largely unexplored. For underactuated systems, reaching unstable equilibria requires explicit exploitation of dynamics, making swing-up tasks \cite{della2020soft, chhatoi2023optimal} canonical benchmarks for dynamic control. 
Additional challenges arise from soft actuation, which imposes intrinsic box constraints on the control inputs due to their unidirectional and distributed nature, which are typically neglected by control methods developed for rigid robots.

\begin{figure}[!t]
\centering
\includegraphics[width=0.95\linewidth]{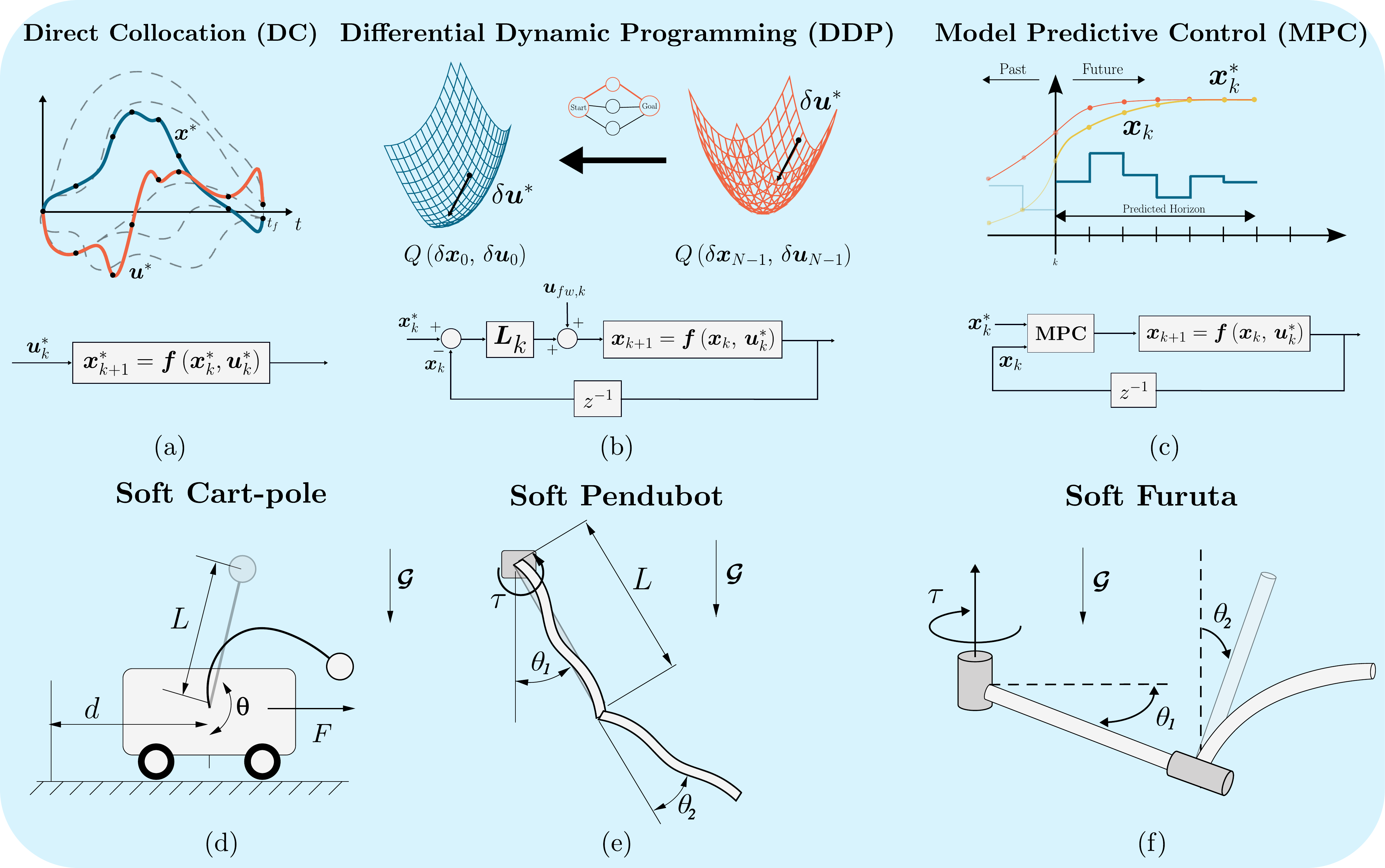}
\caption{Model-based optimal controllers (a--c) applied to soft underactuated systems (d--f). Direct Collocation (a), Differential Dynamic Programming (b), and Model Predictive Control (c) are utilized to perform swing-up tasks on the Soft Cart-Pole (d), Soft Pendubot (e), and Soft Furuta Pendulum (f).}
\label{fig:graphical_abstract}
\end{figure}

Model-based optimal control provides a principled framework to address these challenges by explicitly exploiting system dynamics while handling input constraints. However, its application to dynamic soft robotic systems remains limited, primarily due to the reliance on simplified or low-dimensional models adopted to reduce computational cost. While effective in quasi-static regimes, such approximations are inadequate for highly dynamic maneuvers, where accurate continuum dynamics are essential.

This paper addresses this limitation by leveraging a physically accurate modeling framework based on the \ac{GVS} formulation \cite{mathew2024reduced}, employing variable strain models.
Derived from \ac{CRT}, \ac{GVS} captures all strain modes and complex continuum deformations using a limited number of generalized coordinates, making it well suited for optimal control. Recent advances provide analytical derivatives of the \ac{GVS} dynamics \cite{mathew2024analytical}, enabling efficient gradient-based optimization for high-\ac{DoFs} continuum models.

Building on this framework, we investigate model-based optimal control for dynamic swing-up tasks in rigid-soft underactuated systems. Three model-based optimal control methods—\ac{DC}, \ac{DDP}, and \ac{NMPC}—are evaluated on soft continuum counterparts of classical benchmarks: the Soft Cart-Pole, Soft Pendubot, and Soft Furuta Pendulum (Fig.~\ref{fig:graphical_abstract}). These case studies confirm the framework's capacity to handle non-minimum phase behaviors, complex out-of-plane deformations, and soft-link interactions. All controllers operate on high-\ac{DoFs} continuum models, without quasi-static assumptions or pseudo-rigid approximations.

The dynamics of \acp{CSR} are governed by high-dimensional and stiff differential equations, which necessitate advanced numerical integration schemes~\cite{boyer2023implicit} and handling of input constraints.
To enable the application of \ac{DDP} under these conditions, we combine box-constrained control input with implicit integration by adapting the \ac{Box-DDP}~\cite{tassa2014control} framework within an \ac{IDDP}~\cite{chatzinikolaidis2021trajectory, jallet2022implicit} formulation.
The resulting approach, referred to as \ac{Box-IDDP}, allows efficient optimization of constrained, high-\ac{DoFs} continuum models.
Furthermore, a resolution-based warm-start strategy exploiting the structure of the \ac{GVS} model is further introduced to accelerate the convergence of model-based optimal control methods, thereby enhancing computational efficiency.

\section{Related Works} \label{sec:soa}
In model-based optimal control, selecting a dynamic model requires balancing computational efficiency and physical accuracy. \ac{PCC} models~\cite{webster2010design} are efficient but cannot represent continuous strain, whereas Cosserat rod formulations~\cite{till2019real} and discretization methods such as \ac{DER}~\cite{gazzola2018forward} and DEFORM~\cite{chen2024differentiable} offer high fidelity at the cost of increased dimensionality or challenging analytical derivatives. The \ac{GVS} formulation addresses these issues by parameterizing the strain field, reducing partial differential equations to ordinary differential equations, and extending rigid-body algorithms to rigid–soft systems.

\ac{DC} has been widely applied to \acp{CSR} using \ac{PCC}~\cite{wang2022energy}, pseudo-rigid models~\cite{wertz2022trajectory}, and \ac{FEM} frameworks~\cite{bern2019trajectory, fishman2021control}. High-fidelity \ac{FEM} models exhibit stiff dynamics and contact interactions, typically requiring implicit or semi-implicit integration~\cite{bern2019trajectory, tonkens2021soft, fishman2021control}. Consequently, simplified models are often preferred to reduce computational cost, including \ac{PCC}~\cite{wang2022energy} and pseudo-rigid approximations~\cite{wertz2022trajectory}.

The use of \ac{DDP} for \acp{CSR} remains limited due to the difficulty of deriving analytical derivatives and handling nonlinear continuum dynamics. Early work relied on simplified representations, e.g., \ac{Box-FDDP}~\cite{mastalli2022feasibility} for pseudo-rigid systems~\cite{chhatoi2023optimal} or lumped-mass rods~\cite{pierallini2025fishing}. Recent approaches employ condensed semi-implicit \ac{FEM} models to enable \ac{DDP} on high-\ac{DoFs} systems~\cite{menager2024condensed}, typically using constrained variants such as \ac{Box-DDP}~\cite{tassa2014control}.

\ac{MPC} is the most widely used framework for \acp{CSR} due to its constraint-handling capabilities, though real-time requirements often limit it to simplified or linear models~\cite{best2016new}. To improve accuracy, data-driven approaches based on neural networks~\cite{gillespie2018learning} and Koopman operators~\cite{bruder2019modeling, shi2023koopman, haggerty2023control} have been proposed. Additional strategies include robust~\cite{ouyang2018robust}, adaptive~\cite{hyatt2020model}, and auto-tuned \ac{MPC}~\cite{null2023automatically}. Fully nonlinear \ac{MPC} remains rare and is generally restricted to \ac{PCC} models~\cite{el2020nonlinear, walker2024model}.

This work enables optimal control of rigid–soft systems by exploiting analytical derivatives of the \ac{GVS} model. Unlike pseudo-rigid or quasi-static approaches, it captures full three-dimensional continuum dynamics and enables control around unstable equilibria. The main contributions are: (i) \ac{Box-IDDP}, combining \ac{IDDP}~\cite{jallet2022implicit} with \ac{Box-DDP}~\cite{tassa2014control} for stable, constrained optimization of stiff systems; (ii) a resolution-based warm-start strategy leveraging the \ac{GVS} structure; and (iii) validation on underactuated, non-minimum-phase systems, including the Soft Cart-Pole, Soft Pendubot, and Soft Furuta Pendulum.

\section{Theoretical Background} \label{sec:theoretical_background}
The \ac{GVS} approach \cite{mathew2024reduced} extends classical rigid-body dynamics to hybrid systems, offering a unified \ac{CRT}-based framework for rigid joints and slender soft bodies. The method reduces the governing \textcolor{black}{partial differential equations} to a finite set of \acp{ODE} by parameterizing the strain field. In this work, this is achieved through a truncated expansion of Legendre polynomial basis functions.
A key feature for control applications is the ability to selectively activate individual strain modes (i.e., bending, twisting, stretching, and shearing) and to independently choose the number of \ac{DoFs} associated with each mode.

%
Moreover, \ac{GVS} describes the forward dynamics of a hybrid rigid–soft linkage in the classical Lagrangian form $\bm{M}(\bm{q}) \ddot{\bm{q}} + \bm{h}(\bm{q}, \dot{\bm{q}}) = \bm{B}(\bm{q}) \bm{u}$, where $\bm{q}, \dot{\bm{q}}, \ddot{\bm{q}} \in \mathbb{R}^{n}$ denote the generalized coordinates and their derivatives, and $\bm{u} \in \mathbb{R}^{m}$ is the actuation input. The vector $\bm{q}$ encodes both joint variables and soft-link deformations (bending, twisting, stretching, and shearing). The terms $\bm{M} \in \mathbb{R}^{n \times n}$, $\bm{h} \in \mathbb{R}^{n}$, and $\bm{B} \in \mathbb{R}^{n \times m}$ represent the generalized mass matrix, the vector containing Coriolis, gravitational, stiffness, and damping forces, and the input matrix, respectively. Solving this equation for $\ddot{\bm{q}}$, the forward dynamics can be compactly expressed as
\begin{equation} \label{eq:state-space}
    \begin{bmatrix}
        \dot{\bm{x}}_1 \\ \dot{\bm{x}}_2
    \end{bmatrix} = \begin{bmatrix}
        \bm{x}_2 \\ \textnormal{FD}\left(\bm{x}_1, \, \bm{x}_2, \, \bm{u}\right)
    \end{bmatrix} = \bm{f}\left(\bm{x}, \, \bm{u}\right) \, ,
\end{equation}
where $\bm{x}_1 = \bm{q}$, $\bm{x}_2 = \dot{\bm{q}}$, $\bm{x} = \begin{bmatrix}
    \bm{x}^{\top}_1 & \bm{x}^{\top}_2
\end{bmatrix}^{\top}$, and $\textnormal{FD} = \bm{M}^{-1} \left(\bm{B}\bm{u} - \bm{h}  \right)$.

Equation \eqref{eq:state-space} can be analytically differentiated using the adapted \textcolor{black}{recursive Newton-Euler algorithm} for \ac{GVS} introduced in \cite{mathew2024analytical}, obtaining $\bm{\nabla}_{\bm{q}} \, \textnormal{FD}, \, \bm{\nabla}_{\dot{\bm{q}}} \, \textnormal{FD} \in \mathbb{R}^{n \times n}$.
    
To solve the \ac{TO} problem, the continuous-time dynamics in \eqref{eq:state-space} must be discretized. 
In the literature, time-integration schemes are generally categorized into two classes: explicit and implicit.

Explicit methods compute the next state $\bm{x}_{k+1}$ directly as a function $\bm{f}_d$ of the current state $\bm{x}_k$ and input $\bm{u}_k$, i.e., $\bm{x}_{k+1} = \bm{f}_d(\bm{x}_k, \bm{u}_k)$. In contrast, implicit methods are formalized as
\begin{equation} \label{eq:implicit_time}
    \bm{g}\left(\bm{x}_{k+1}, \bm{x}_k, \bm{u}_{k}\right) = \bm{0} \, ,
\end{equation}
where $\bm{g}(\cdot)$ is the method-specific residual function.
Unlike explicit schemes, \eqref{eq:implicit_time} generally cannot be solved in closed form for nonlinear systems. Therefore, iterative techniques are required, such as the Newton-Raphson method~\cite[Chap.~5]{GallardoAlvarado2022}.

\color{black}
High-order \ac{CSR} models are often governed by stiff \acp{ODE}, making implicit integration necessary for stable simulation and control~\cite{boyer2023implicit}. Unlike explicit solvers, which require extremely small time steps to avoid divergence, implicit methods remain stable with larger time steps. Although each step is computationally more demanding, implicit schemes are generally more suitable for real-time optimal control, especially when analytical derivatives are available to improve the efficiency and convergence of iterative solvers.
\color{black}

\section{Model-based Optimal Controllers} \label{sec:mb_opt}
Given the dynamic model~\eqref{eq:state-space} and the initial condition $\bm{x}_0$, the goal of the \ac{TO} problem is to compute an optimal policy $\bm{u}^*(t)$ that solves
\begin{equation}\label{eq:continous_ocp}
    \begin{aligned}
        \min_{\bm{u}} \quad &\ell_{f}\left(\bm{x}\left(t_f\right)\right) + \int_{0}^{t_f} \ell\left(\bm{x}\left(t\right), \bm{u}\left(t\right)\right) \, \textnormal{d}t \\
        \textnormal{s.t.} \quad & \dot{\bm{x}} = \bm{f}\left(\bm{x}\left(t\right), \bm{u}\left(t\right)\right) \quad t \in [0, \, t_f] \\
        &\bm{u}_{\textnormal{lb}} \leq \bm{u}_k \leq \bm{u}_{\textnormal{ub}} \quad t \in [0, \, t_f]
    \end{aligned} \, ,
\end{equation}
where $\ell : \mathbb{R}^{2n} \times \mathbb{R}^{m} \rightarrow \mathbb{R}^{+}$ is the cost, and $\ell_f : \mathbb{R}^{2n} \rightarrow \mathbb{R}^{+}$ is the terminal cost. Furthermore, the vectors $\bm{u}_{\textnormal{lb}}, \bm{u}_{\textnormal{ub}} \in \mathbb{R}^{m}$ define lower and upper bounds on the control input, commonly referred to as box constraints.
The \ac{OCP} in \eqref{eq:continous_ocp} can generally be further constrained by imposing nonlinear inequality or equality constraints on the state and input.

To formalize~\eqref{eq:continous_ocp} as a numerical optimization problem, the \ac{OCP} is discretized using two main approaches: (i) direct transcription and (ii) direct shooting.
In direct transcription, the discrete-time states $\bm{x}_k$ and inputs $\bm{u}_k$ are treated as decision variables, leading to a large but sparse \ac{NLP} in which the system dynamics are enforced as constraints at each time step.
In contrast, direct shooting considers only the input trajectory as decision variables, while the state trajectory is obtained by integration of the discretized dynamics.

\subsection{\acf{DC}} \label{mb_opt:dc}
    To reduce the number of decision variables in the direct transcription of~\eqref{eq:continous_ocp}, \ac{DC}~\cite{hargraves1987direct} approximates the state and control trajectories using piecewise polynomial functions, while enforcing the system dynamics as algebraic constraints at a finite set of collocation points.

In this work, we adopt the trapezoidal collocation scheme, which leads to the following \ac{NLP} formulation
\begin{equation}\label{eq:direct_collocation}
    \begin{split}
        \min_{\bm{x}, \bm{u}} \quad &\ell_f\left(\bm{x}_{N}\right) + \sum_{k = 0}^{N - 1} \frac{h}{2} \left(\ell_{k} + \ell_{k + 1}\right) \\
        \textnormal{s.t.} \quad &\bm{x}_{k + 1} = \bm{x}_{k} + \frac{h}{2} \left(\bm{f}_{k} + \bm{f}_{k + 1}\right) \quad k \in [0, N - 1] \\
        & \bm{u}_{\textnormal{lb}} \leq \bm{u}_k \leq \bm{u}_{\textnormal{ub}} \quad k \in [0, N - 1]
    \end{split} \, ,
\end{equation}
Here, $h$ denotes the integration time step, and the collocation points are located at the midpoints between two consecutive time instants $t_k$ and $t_{k+1}$. 
The resulting \ac{NLP} in \eqref{eq:direct_collocation} can be solved using standard optimization methods such as \textcolor{black}{interior-point} or \textcolor{black}{sequential quadratic programming}.
Given the complexity of the high-order rigid-soft models, we selected the \textcolor{black}{interior-point} algorithm due to its suitability for large-scale problems. Additionally, analytical gradients were employed to enhance both solution accuracy and computational efficiency.

The main advantage of direct collocation lies in its simplicity and its ability to naturally handle nonlinear constraints on both inputs and states, making it well-suited for offline trajectory planning.
However, \ac{DC} produces an optimal open-loop solution $\{\bm{x}^*, \bm{u}^*\}$, and thus does not inherently provide robustness to model uncertainties or external disturbances. 
As a result, convergence to the desired trajectory $\bm{x}^*$ is not guaranteed in closed-loop execution. 
Moreover, despite the reduction in decision variables compared to full direct transcription, the computational burden of \ac{DC} grows rapidly with the number of \ac{DoFs}, limiting its applicability to low-dimensional systems.

\subsection{\acf{Box-IDDP}} \label{mb_opt:ddp}
    By decomposing the original problem into a sequence of smaller subproblems, \ac{DDP}~\cite{mayne1966second,jacobson1968differential} solves the \ac{OCP} by recursively applying Bellman’s principle of optimality backward in time.

Unlike \ac{DC}, which formulates the problem as a large-scale \ac{NLP}, \ac{DDP} employs a direct shooting approach that iteratively improves the solution through two main phases: a backward pass and a forward pass.
In the backward pass, the algorithm constructs quadratic approximations of the cost-to-go function to compute local optimal control policies. 
In the forward pass, these policies are applied to the nonlinear system dynamics to generate an updated trajectory.

Classical \ac{DDP}, however, assumes explicit time discretization and does not account for box constraints on the control inputs. These assumptions are not suitable for \acp{CSR}, which typically require implicit time integration and bounded actuation. To address these limitations, we combine \ac{Box-DDP}~\cite{tassa2014control}, which enforces box input constraints, with \ac{IDDP}~\cite{chatzinikolaidis2021trajectory,jallet2022implicit}, which extends DDP to implicitly discretized dynamics. We refer to the resulting method as \ac{Box-IDDP}.

The backward pass begins with the Bellman equation at time step $k$, given by \textcolor{black}{$V\left(\bm{x}_k\right)
    =
    \min_{\bm{u}_k}
    \left(
        \ell\left(\bm{x}_k, \bm{u}_k\right)
        +
        V\left(\bm{x}_{k+1}\right)
    \right)$},
where $V(\cdot)$ denotes the value function.

Let $Q\left(\delta \bm{x}_k, \delta \bm{u}_k\right)$ be the local variation of
the argument of the minimization in \textcolor{black}{the previous equation}, expressed in terms
of perturbations around the nominal trajectory.
A second-order Taylor expansion of $Q$ yields
\begin{equation}
\label{eq:taylor_action}
    Q\left(\delta \bm{x}, \delta \bm{u}\right)
    \approx
    \frac{1}{2}
    \begin{bmatrix}
        1 \\
        \delta \bm{x} \\
        \delta \bm{u}
    \end{bmatrix}^{\top}
    \begin{bmatrix}
        0 & \bm{Q}_x^{\top} & \bm{Q}_u^{\top} \\
        \bm{Q}_x & \bm{Q}_{xx} & \bm{Q}_{xu} \\
        \bm{Q}_u & \bm{Q}_{ux} & \bm{Q}_{uu}
    \end{bmatrix}
    \begin{bmatrix}
        1 \\
        \delta \bm{x} \\
        \delta \bm{u}
    \end{bmatrix},
\end{equation}
where the matrices $\bm{Q}_{(\cdot)}$ collect the first- and second-order
derivatives of the cost and dynamics with respect to the state and control.

To accommodate implicit time integration, \ac{IDDP} modifies the computation of
the $Q$-function derivatives to account for the implicit integration scheme in
\eqref{eq:implicit_time}, \textcolor{black}{reported in \eqref{eq:Qx_Qu}-\eqref{eq:Vxx}}.

To enforce box constraints on the control inputs, \ac{Box-DDP} computes the locally
optimal control variation by solving a constrained quadratic program.
Specifically, the backward pass computes
\begin{equation}
\label{eq:backward_pass}
\begin{aligned}
    \delta \bm{u}_k = &\arg \min_{\delta \bm{u}_k} Q\left(\delta \bm{x}_k, \delta \bm{u}_k\right) = \bm{l}_k + \bm{L}_k \, \delta \bm{x}_k \\
    &\text{s.t.} \quad
    \bm{u}_{\mathrm{lb}} \leq \bm{u}_k + \delta \bm{u}_k \leq \bm{u}_{\mathrm{ub}}
\end{aligned} \, ,
\end{equation}
where the feedforward term $\bm{l}_k \in \mathbb{R}^{m}$ and the feedback gain $\bm{L}_k \in \mathbb{R}^{m \times 2n}$ are
computed using the Box-QP algorithm~\cite{tassa2014control}.
This procedure is applied recursively from the terminal time step to the initial one.

Consequently, during the forward pass, the system is rolled out using a line search on the feedforward term $\bm{l}_k$, scaled by a factor $\alpha \in (0,1]$, while integrating the dynamics using the implicit scheme~\eqref{eq:implicit_time}.

At convergence, the method returns a nominal trajectory
$\{\bm{x}^*, \bm{u}^*\}$ together with a time-varying linear feedback law
$\bm{u}_k = \bm{u}_k^{*} + \bm{L}_k \left( \bm{x}_k^{*} - \bm{x}_k \right)$, which enables local trajectory tracking within the region of validity of the
second-order approximation.

Compared to \ac{DC}, \ac{DDP} directly yields a locally optimal feedback controller
with significantly lower computational complexity, since it optimizes only over
the control sequence.
However, due to its reliance on local approximations of the cost and dynamics,
\ac{DDP} is highly sensitive to the initial guess, making effective warm-start
strategies essential for practical applications.

\subsection{\acf{NMPC}} \label{mb_opt:nmpc}
    \ac{NMPC}~\cite{grne2013nonlinear} solves the \ac{TO} problem online over a finite prediction horizon using state feedback. At each time step, only the first control input of the optimized sequence is applied to the system, and the optimization problem is subsequently resolved in a receding-horizon fashion, warm-started from the previously computed solution.

At each time step $k$, given the feedback $\bm{x}_k$, the \ac{NMPC} solves the following finite-horizon \ac{TO} problem
\begin{equation}\label{eq:nlmpc_formulation}
    \begin{aligned}
        \min_{\substack{
            \bm{x}_{k+1}, \dots, \bm{x}_{k + p + 1},\\
            \bm{u}_{k}, \dots, \bm{u}_{k + p}
        }}
        \quad &
        \ell_f\left(\bm{x}_{k + p + 1}\right)
        + \sum_{i = k}^{k + p} \ell\left(\bm{x}_i, \bm{u}_i\right) \\
        \textnormal{s.t.} \quad &
        \bm{g}\left(\bm{x}_i, \bm{x}_{i + 1}, \bm{u}_{i}\right) = \bm{0}
        \quad i \in [k, \, k + p] \\
        &
        \bm{u}_{\textnormal{lb}} \leq \bm{u}_i \leq \bm{u}_{\textnormal{ub}}
        \quad i \in [k, \, k + p]
    \end{aligned} \, ,
\end{equation}
Here, $p \in \mathbb{N}$ denotes the prediction horizon length. The optimization problem in \eqref{eq:nlmpc_formulation} can be solved using either \ac{DC} or \ac{DDP}-based methods.

By exploiting state feedback, \ac{NMPC} provides robustness to disturbances and model uncertainties. 
However, its closed-loop performance and convergence properties are strongly influenced by the choice of hyperparameters, such as the prediction horizon $p$, and by the quality of the initial guess used to warm-start the solver.

In this work, we implement \ac{NMPC} by solving the underlying \ac{TO} problem using \ac{Box-IDDP}, leveraging its computational efficiency and faster convergence compared to \ac{DC}, particularly when a suitable warm-start is available.

\subsection{Warm-start Strategy} \label{mb_opt:warm_start}
    \color{black}
To improve the convergence rate and computational efficiency of optimal control algorithms, we propose a warm-start strategy for hybrid rigid–soft robots modeled using the \ac{GVS} approach. The main idea is to solve the \ac{TO} problem on progressively higher-order models, starting from a reduced-order representation and increasing the number of \ac{DoFs} until the desired truncation is reached. 
Since higher-order terms in the \ac{GVS} strain expansion~\cite{mathew2024reduced} have a progressively smaller influence on the system dynamics, solutions obtained with lower-order models provide effective initial guesses for higher-order formulations.

\begin{figure}[!t]
    \centering
    \includegraphics[width=0.8\linewidth]{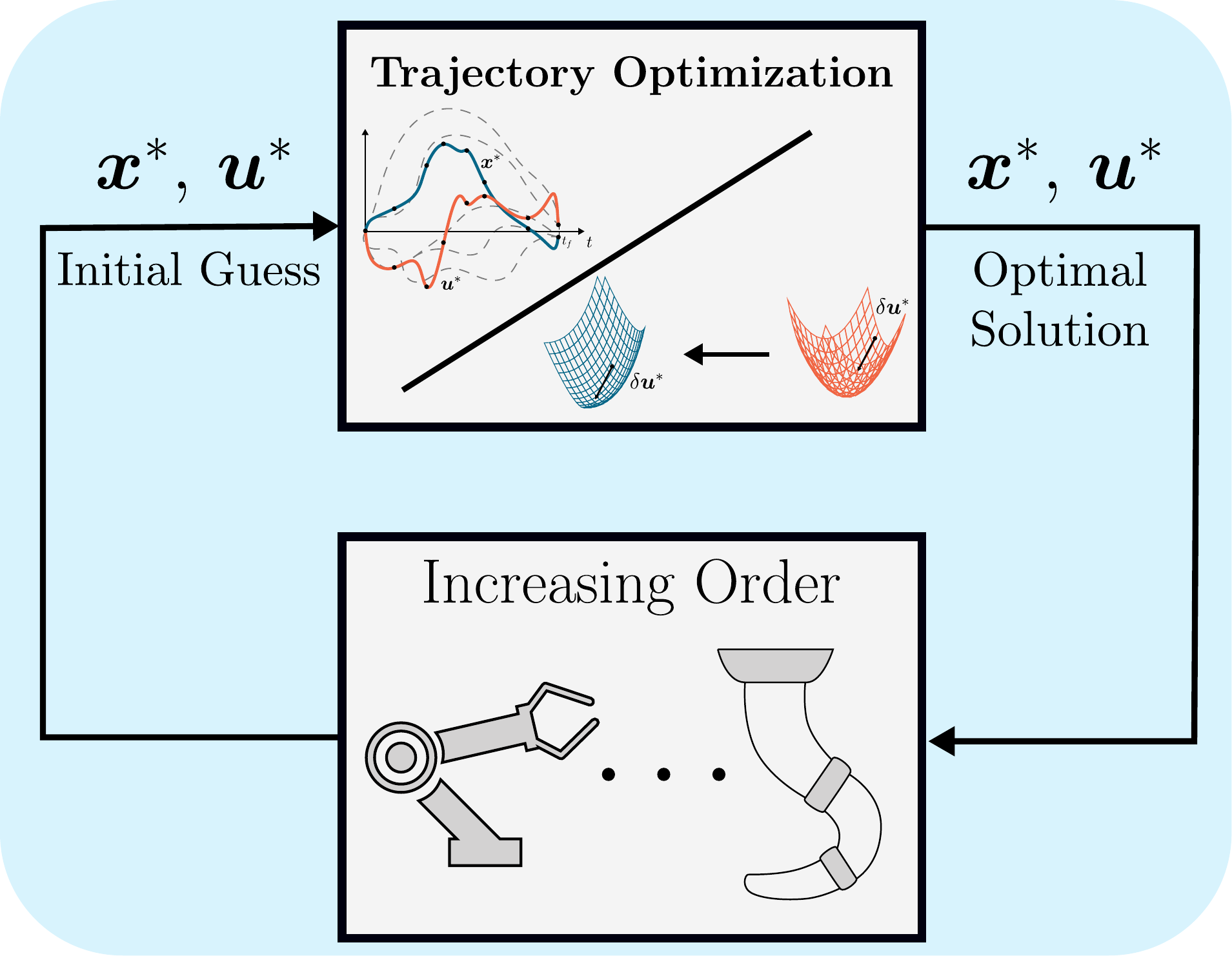}
    \caption{Representative scheme of the proposed warm-start strategy. The initial guess for the optimal controllers is obtained by solving simplified versions of the system, starting from the rigid equivalent model and then gradually adding the \ac{DoFs}.}
    \label{fig:warm-start}
\end{figure}

This strategy exploits the strain-based parameterization of \ac{GVS}, which allows selectively enabling specific strain modes. Such a feature is intrinsic to the formulation and is not available in alternative approaches such as \ac{DER}~\cite{gazzola2018forward}. 
Fig.~\ref{fig:warm-start} illustrates the proposed procedure. 
First, the \ac{TO} problem is solved for the equivalent rigid system. The solution is then transferred to a soft model with a constant-curvature parameterization, where only the curvature mode is enabled and assumed constant along the link. This represents the simplest soft formulation and provides a smooth transition from the rigid case, as the constant-curvature assumption constrains the link to circular-arc deformations, effectively making it stiffer.

The model complexity is then increased by progressively enabling additional curvature \ac{DoFs}. In this work, the curvature is expanded up to the second-order Legendre polynomial basis. Finally, linear strain modes are introduced at the desired resolution, using the previous solution as initialization; similarly, second-order expansions are adopted. This gradual refinement is motivated by the fact that bending is typically the dominant deformation mode in rod-like bodies, while stretch and shear have a smaller but non-negligible influence, especially during dynamic interactions.

Although the complete warm-start procedure introduces additional computational cost compared with random initialization, random guesses often lead to solver failures or convergence to poor local minima, particularly for complex underactuated systems. Therefore, the proposed approach is especially suitable for offline \acs{TO}. In \acs{NMPC}, it can initialize the first optimization problem, improving convergence toward the desired trajectory without affecting subsequent real-time performance.
\color{black}

\section{Case Studies on Rigid-Soft Systems} \label{sec:case_studies}
    The model-based optimal controllers are evaluated on three high-order systems of increasing complexity. These range from the Soft Cart-Pole, with a single planar soft link, to the Soft Pendubot, composed of two planar soft links, and finally the Soft Furuta Pendulum—a full three-dimensional hybrid system exhibiting all strain modes.

All systems are modeled using the \ac{GVS} approach within the Differentiable \ac{SoRoSim} framework~\cite{mathew2022sorosim, mathew2024analytical}. The soft links are cylindrical with length $L = \SI{1.0}{m}$, cross-section radius $R_\textnormal{cs} = \SI{0.03}{m}$, density $\bar{\rho} = \SI{1000}{kg/m^3}$, Young's modulus $E = \SI{1.0}{MPa}$, Poisson ratio $\nu = 0.5$, damping $\beta = \SI{0.01}{MPa\cdot s}$, stress-free strain $\bm{\xi}^* = [0, 0, 0, 1, 0, 0]^\top$, and joint damping $\beta_r = \SI{0.05}{\newton\meter s/rad}$.

Each soft link is parameterized via second-order Legendre polynomials, yielding 3 \ac{DoFs} for each strain mode.
The target configuration $\bar{\bm{q}}$ corresponds to the unstable swing-up equilibrium, with target state $\bm{x}_{\textnormal{target}} = [\bar{\bm{q}}^\top, \bm{0}^\top]^\top \in \mathbb{R}^{2n}$.

Swing-up control uses a quadratic cost $\ell = \frac{1}{2}(\bm{e}^\top \bm{Q} \bm{e} + \bm{u}^\top \bm{R} \bm{u})$, $\ell_f = \frac{1}{2} \bm{e}_{f}^{\top} \bm{Q}_f \bm{e}_{f}$, with state error $\bm{e} = \bm{x}_{\textnormal{target}} - \bm{x}$ and box-constrained inputs $-u_{\max} \leq u \leq u_{\max}$. Weights $\left(\bm{Q}_{f}, \bm{Q}, \bm{R}\right)$ are identical across methods.

Concerning time discretization, \ac{DC}~\cite{kelly2017introduction} utilizes trapezoidal integration with $N = 100$ nodes over $t_f = \SI{2}{\second}$, whereas \ac{Box-IDDP} and \ac{NMPC} employ implicit Euler integration with a step size of $h = \SI{0.01}{\second}$. The latter methods solve~\eqref{eq:implicit_time} using the trust-region dogleg algorithm~\cite{coleman1996interior}, leveraging the analytical derivatives of~\eqref{eq:state-space}. Additionally, \ac{NMPC} uses a prediction horizon of \SI{1.0}{\second} ($p = 100$).
The warm-start strategy is implemented by sequentially solving the reduced-order parameterizations detailed in Sec.~\ref{mb_opt:warm_start}.

\begin{figure*}[!t]
    \centering
    \includegraphics[width=0.95\linewidth]{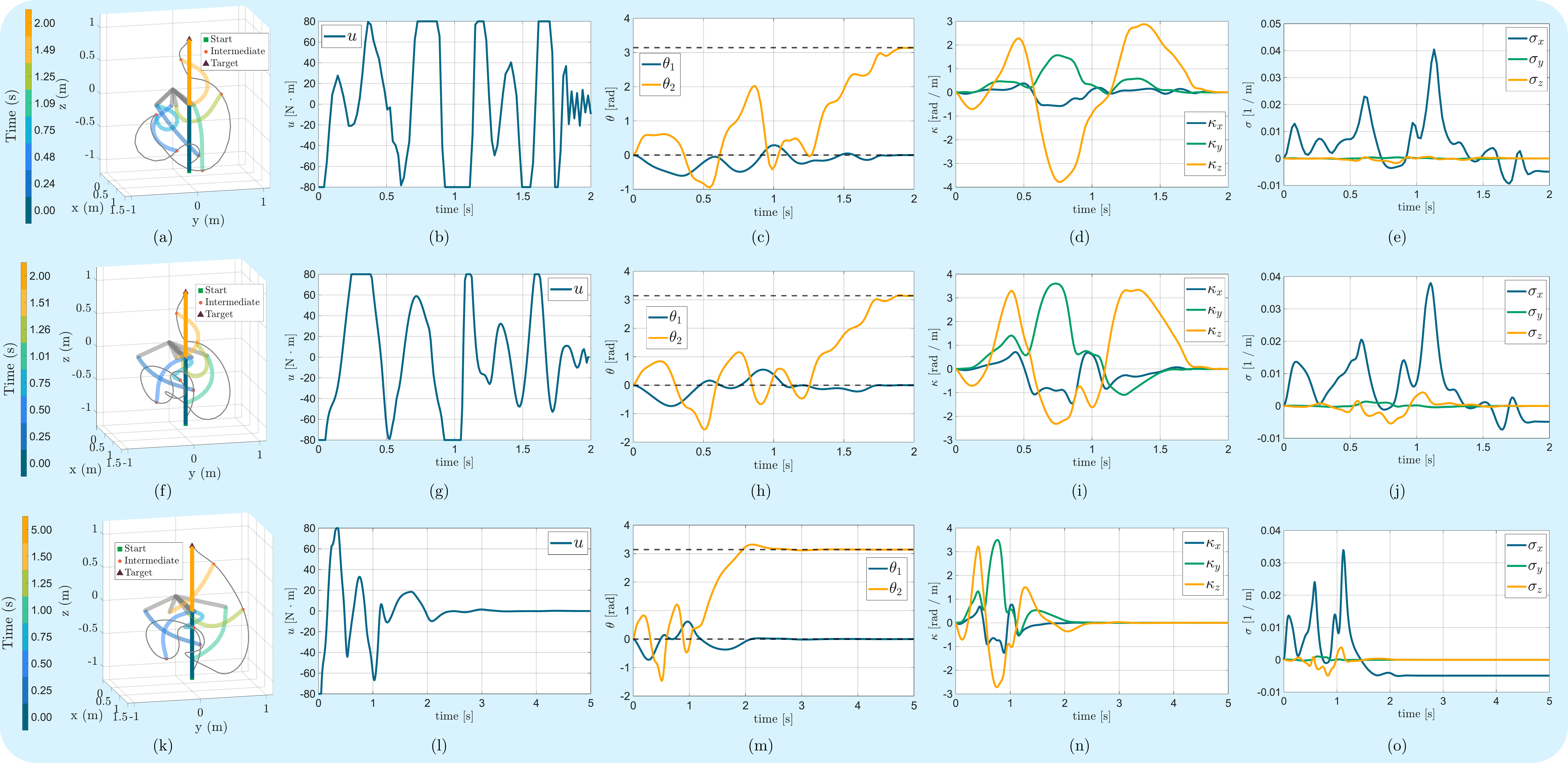}
    \caption{Swing-up Task using \ac{DC} (a-e), \ac{Box-IDDP} (f-j), and \ac{NMPC} (k-o) on the Soft Furuta pendulum.}
    \label{fig:soft_furuta3D}
\end{figure*}

Figs. \ref{fig:soft_cartpole}--\ref{fig:soft_furuta3D} illustrate the swing-up tasks across the different systems. For each optimal controller, we show the control inputs and joint trajectories. 
To better isolate the strain modes, we report the spatially averaged \textcolor{black}{angular strain $\bm{\kappa} \in \mathbb{R}^{3}$ and average linear strain $\bm{\sigma} \in \mathbb{R}^{3}$} of the soft links rather than the full configuration vector.
Furthermore, for brevity, only the final results for the high-order systems are shown.

\subsection{Soft Cart-pole} \label{case_studies:cs1}
    \begin{figure*}[!t]
    \centering
    \includegraphics[width=0.95\linewidth]{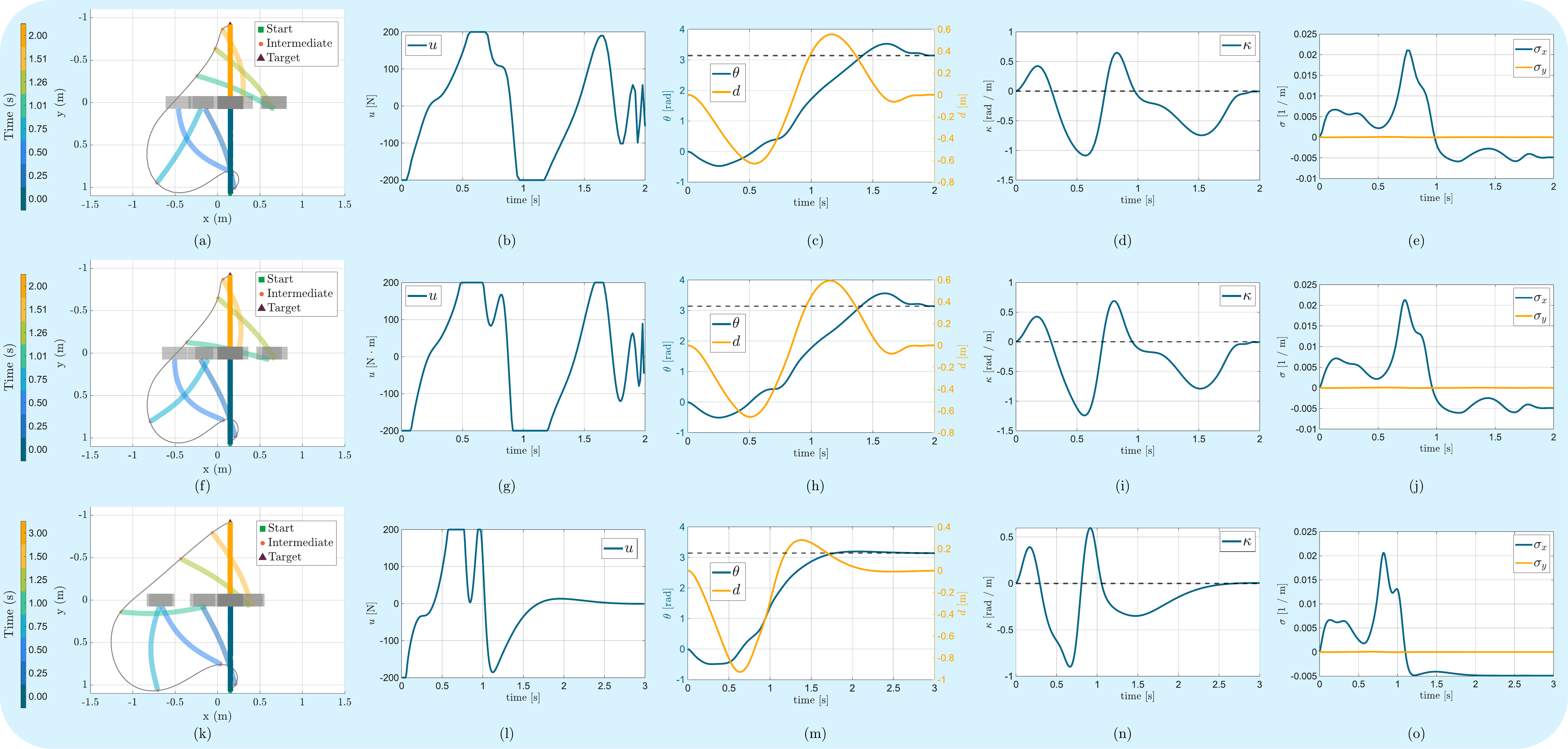}
    \caption{Swing-up Task using \ac{DC} (a-e), \ac{Box-IDDP} (f-j), and \ac{NMPC} (k-o) on the Soft Cart-pole.}
    \label{fig:soft_cartpole}
\end{figure*}
The Soft Cart-Pole (Fig.~\ref{fig:graphical_abstract}d) consists of a rigid cart that translates along a single axis under the action of an external force $F(t) \in \mathbb{R}$, producing a displacement $d(t) \in \mathbb{R}$. A soft link is connected to the cart via a revolute joint with angle $\theta(t) \in \mathbb{R}$, subject to rotational damping $\beta_r$. The configuration vector is defined as $\bm{q} =
\begin{bmatrix}
    d &
    \theta &
    \bm{q}_{\bm{\xi}}^{\top}
\end{bmatrix}^{\top} \in \mathbb{R}^{n}$, where $\bm{q}_{\xi}$ denotes the configuration of the soft link, and $n = 11$.
The swing-up task is achieved when the soft link reaches the upright and straight configuration, corresponding to $\theta = \pi$. The target equilibrium $\bar{\bm{q}}$ is defined at $d = 0$, corresponding to the static upright equilibrium of the soft link. Moreover, the input is subject to a box constraint with $F_{\max} = \SI{200}{\newton}$.

Fig.~\ref{fig:soft_cartpole} illustrates the application of \ac{DC} (a–e), \ac{Box-IDDP} (f–j), and \ac{NMPC} (k–o) to the Soft Cart-pole. All methods produce optimal policies that successfully drive the system to perform the swing-up maneuver.
\color{black}
The norm of the final error, $\|\bm{e}_f\|$, is $0.0032$ for \ac{DC}, $0.0048$ for \ac{Box-IDDP}, and $0.0191$ for \ac{NMPC}.
\color{black}

Due to input saturation, the controllers exhibit characteristic non-minimum phase behavior, driving the cart in the opposite direction to accumulate energy. This strategy leverages the combined effects of inertia and elasticity to propel the soft link to the upright configuration.

 Concerning the \ac{DC} and \ac{Box-IDDP} controllers, the final phase of the optimal policy consists of rapid oscillatory inputs to counteract the gravity-induced bending of the soft link due to its own weight.
 Conversely, \ac{NMPC} exploits a larger cart displacement, allowing the system to converge smoothly to the target, at the cost of a slower settling time.

\subsection{Soft Pendubot} \label{case_studies:cs2}
    \begin{figure*}[!t]
    \centering
    \includegraphics[width=0.95\linewidth]{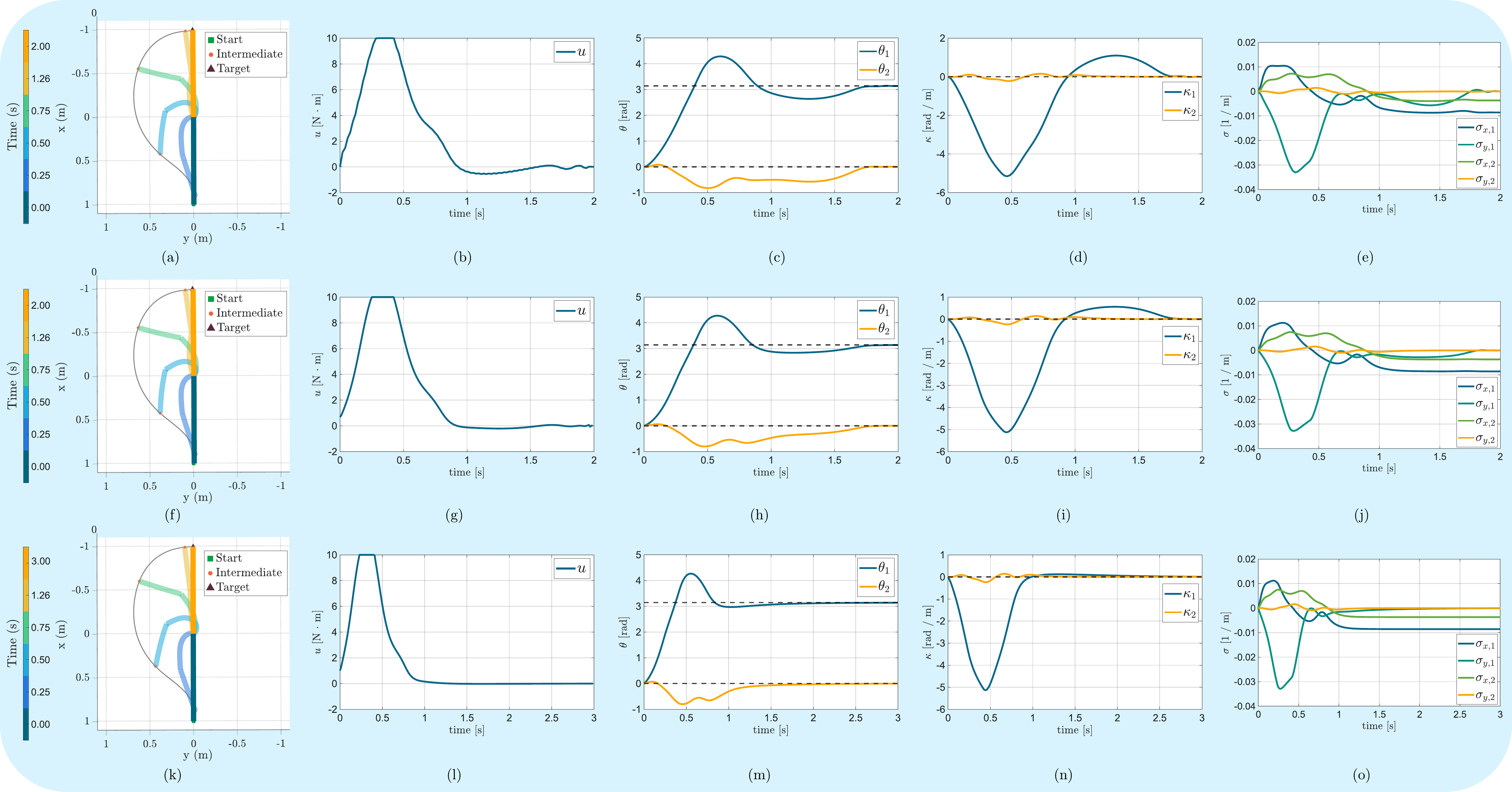}
    \caption{Swing-up Task using \ac{DC} (a-e), \ac{Box-IDDP} (f-j), and \ac{NMPC} (k-o) on the Soft Pendubot.}
    \label{fig:soft_pendubot}
\end{figure*}
The Soft Pendubot (Fig.~\ref{fig:graphical_abstract}e) is an open kinematic chain composed of two soft links, each of length $L/2$. These are connected by revolute joints $\theta_1 \in \mathbb{R}$ and $\theta_2 \in \mathbb{R}$\textcolor{black}{, subjected to rotational damping $\beta_r$}.
The first joint, located at the base, is actuated by a torque input $\tau \in \mathbb{R}$, with $\tau_{\max} = \SI{10}{\newton}$, while the second joint is passive. The full configuration vector of the Soft Pendubot is defined as $\bm{q} =
\begin{bmatrix}
    \theta_1 &
    \bm{q}_{\bm{\xi},1}^{\top} &
    \theta_2 &
    \bm{q}_{\bm{\xi},2}^{\top}
\end{bmatrix}^{\top} \in \mathbb{R}^{n}$, with $n = 20$. The Swing-up configuration correspond to the unstable equilibrium for $\theta_1 = \pi$ and $\theta_2 = 0$.

Fig.~\ref{fig:soft_pendubot} illustrates the application of \ac{DC} (a–e), \ac{Box-IDDP} (f–j), and \ac{NMPC} (k–o) to the Soft Pendubot. In this case study, the input box constraints allow the robot to be driven directly to the upright configuration without evident non-minimum phase behavior.
\color{black}
The norm of the final error, $\|\bm{e}_f\|$, is $0.0075$ for \ac{DC}, $0.0029$ for \ac{Box-IDDP}, and $0.027$ for \ac{NMPC}.
\color{black}

Notably, the coupling between the links becomes significant as the robot approaches the upright configuration. The first link undergoes gravity-induced bending due to the combined load of its own mass and the second link. The optimal policies counter this deformation via a negative torque, acting to maintain the upright configuration and prevent the system from reverting to the rest configuration.

\subsection{Soft Furuta Pendulum} \label{case_studies:cs3}
The Soft Furuta pendulum (Fig.~\ref{fig:graphical_abstract}f) consists of a rigid link attached to a revolute joint $\theta_1 \in \mathbb{R}$, with a soft link connected through a second revolute joint $\theta_2 \in \mathbb{R}$ whose rotation axis is perpendicular to that of the first joint.
This configuration allows the soft link to swing freely while the rigid arm rotates in the horizontal plane. Furthermore, the soft link can exert all strain modes, including out-of-plane bending, shear, and twisting.
The configuration vector is defined as $\bm{q} =
\begin{bmatrix}
    \theta_1 &
    \theta_2 &
    \bm{q}_{\bm{\xi}}^{\top}
\end{bmatrix}^{\top} \in \mathbb{R}^{n}$, where $n = 20$.
\color{black}
Similar to the previous case, both revolute joints are subjected to the rotational damping $\beta_r$.
\color{black}
The target equilibrium $\bar{\bm{q}}$ is defined with $\theta_2 = \pi$ and for $\theta_1 = 0$ and the associated static equilibrium of the soft link. The input is subject to a box constraint with $\tau_{\max} = \SI{80}{\newton}$.
\color{black}
The norm of the final error, $\|\bm{e}_f\|$, is $0.0049$ for \ac{DC}, $0.0030$ for \ac{Box-IDDP}, and $0.0055$ for \ac{NMPC}.
\color{black}

Similar to the Soft Cart-Pole (Sec.\ref{case_studies:cs1}), due to input saturation, the optimal policies exhibit non-minimum-phase behavior, oscillating the rigid link to accumulate energy that drives the soft link to the upright configuration. Furthermore, in this case, the out-of-plane strain acts as a disturbance, and the optimal policy must take it into account.

Unlike the previous cases, the three methods differ in the optimal strategies to perform the task.
The \ac{Box-IDDP} solution causes strong out-of-plane bending ($\kappa_y$) and torsion ($\kappa_x$), effectively winding the soft link into a helicoidal shape. The controller leverages the elastic energy stored during this deformation, releasing it to generate the necessary kinetic energy for the task. Once this combined bending and twisting is released, the motion is primarily characterized by in-plane bending ($\kappa_z$), guiding $\theta_2$ to the upright position.
In contrast, the \ac{DC} solution excites the out-of-plane bending $\kappa_y$ significantly less. Consequently, the aforementioned helicoidal effect is less pronounced, and the controller relies primarily on in-plane bending $\kappa_z$.
Finally, as the \ac{NMPC} relies on \ac{Box-IDDP}, it inherits a similar winding strategy to accumulate energy before driving the link to the upright equilibrium. Notably, the \ac{NMPC} achieves asymptotic stabilization with significantly smoother control trajectories, albeit with a slower settling time.
\vspace{-7pt}
\begin{figure}[!t]
    \centering
    \includegraphics[width=0.95\linewidth]{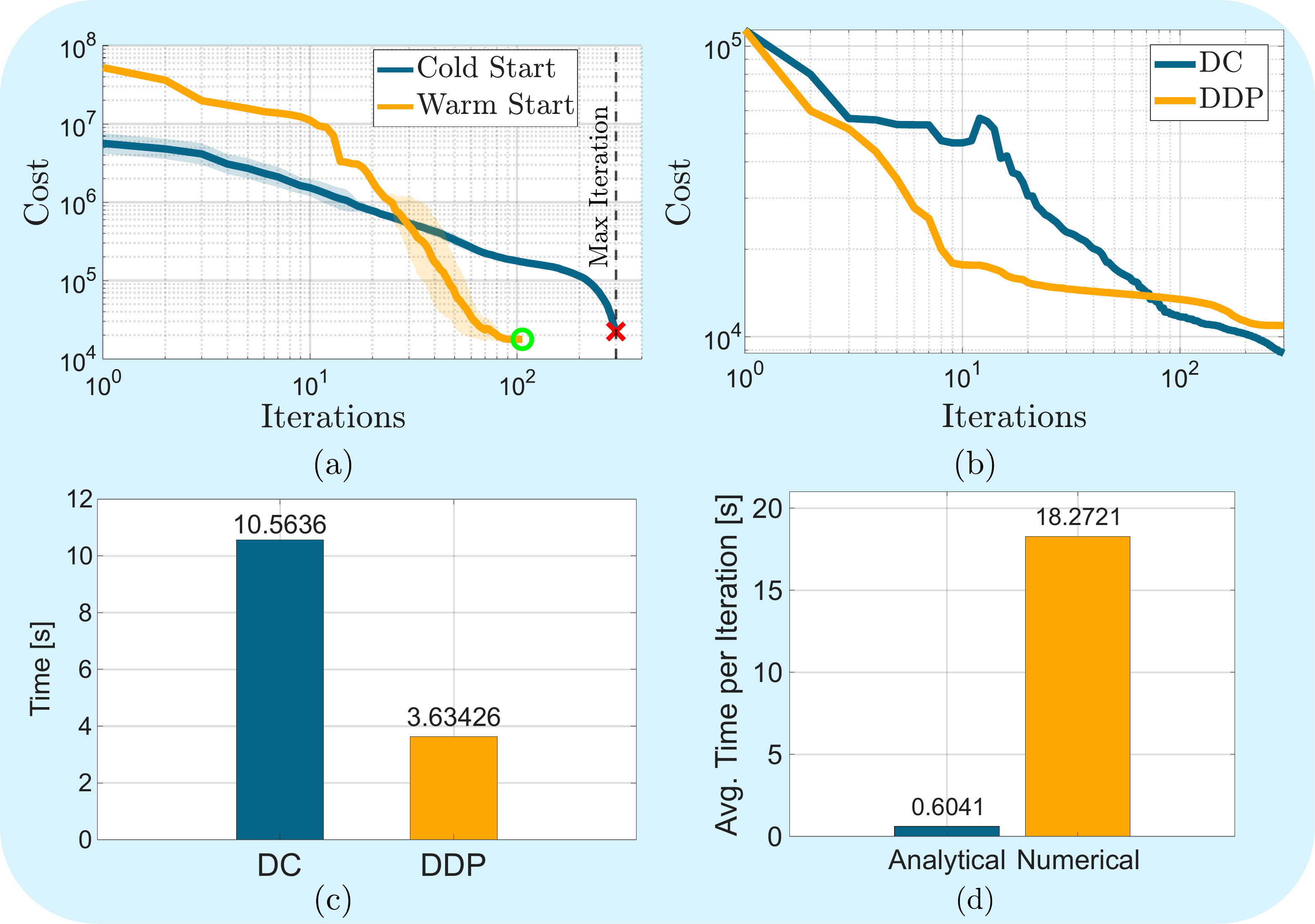}
    \caption{Benefits of the warm-start strategy and performance comparison. (a) Cost over iterations for the Soft \textcolor{black}{Cart-Pole} using \ac{Box-IDDP}, comparing \textcolor{black}{cold start (blue)} vs. warm-start \textcolor{black}{(yellow)}. (b–c) Comparison between \ac{DC} and \ac{Box-IDDP} on the Soft Cart-Pole: (b) cost convergence using identical pseudo-random initial guesses, and (c) average computational time per iteration. \textcolor{black}{(d) Comparison of the average computational time per iteration using analytical versus numerical gradients within the \ac{Box-IDDP} framework.}}
    \label{fig:ws_dc_ddp}
\end{figure}
\subsection{Discussions}
The case studies validate the proposed method across high-order hybrid systems of increasing complexity. The Soft Cart-Pole highlights non-minimum phase behavior under input constraints, while the Soft Pendubot demonstrates policy adaptation to dynamic interactions between soft links. Finally, the Soft Furuta Pendulum extends this to a three-dimensional system, exploiting complex out-of-plane deformations for swing-up maneuvers. This final case also reinforces the non-minimum phase characteristics seen in Sec.~\ref{case_studies:cs1}, requiring strategic energy accumulation to reach the upright configuration.

\subsubsection{Benefits of the Warm-start Strategy}
\color{black}
We evaluated the proposed warm-start strategy using \ac{Box-IDDP}, the algorithm most sensitive to the initial guess among the considered methods.
Fig.~\ref{fig:ws_dc_ddp}a compares the optimization cost obtained with a cold start (10 pseudo-random input sequences) and the warm-start strategy of Sec.~\ref{mb_opt:warm_start}. In both cases, the rigid-model optimization is initialized with the same pseudo-random input sequence. The curves show the mean cost over all trials, and the shaded regions denote one standard deviation.
Although the warm-start exhibits a higher initial cost, it converges approximately 200 iterations earlier. In contrast, the cold start reaches the maximum number of iterations without satisfying the convergence criterion.
\color{black}

\color{black}
\subsubsection{Comparison between \ac{DC} and \ac{Box-IDDP}}
We further compared \ac{DC} and \ac{Box-IDDP} on the Soft Cart-Pole without warm-start, using identical pseudo-random initial guesses and the same optimization hyperparameters adopted in the case studies.
Fig.~\ref{fig:ws_dc_ddp}b--c report the cost evolution and the average computational time per iteration. Simulations were performed in MATLAB 2025a on a laptop with an Intel Core i7-12700H processor (2.30 GHz) and 16 GB of RAM.
\ac{Box-IDDP} reduces the cost more rapidly than \ac{DC}, whose cost temporarily increases for satisfying constraints within the large \ac{NLP}. However, \ac{Box-IDDP} converges to a suboptimal solution with a higher final cost. In terms of computational efficiency, \ac{DC} requires 2.9$\times$ more computation per iteration.
Overall, \ac{Box-IDDP} offers faster convergence and lower computational cost, making it better suited for real-time applications such as \ac{NMPC}. In contrast, \ac{DC} remains preferable for offline planning problems involving complex state and input constraints, such as collision avoidance.
\color{black}

\color{black}
\subsubsection{Benefits of Analytical Gradients}
To quantify the computational benefits of differentiable physics, we compared analytical and numerical (central-difference) gradients within \ac{Box-IDDP} on the Soft Cart-Pole. Over 10 runs with different pseudo-random initial guesses, analytical gradients reduced the average iteration time from \SI{18.2721}{\second} to \SI{0.6041}{\second} (Fig.~\ref{fig:ws_dc_ddp}d), demonstrating a substantial computational advantage over numerical differentiation.
\color{black}

\section{Conclusion} \label{sec:conclusion}
This paper presented a model-based optimal control framework for underactuated rigid-soft robots using the \ac{GVS} approach. This formulation enables efficient gradient-based optimization on full continuum dynamics, avoiding quasi-static or pseudo-rigid approximations.

We validated \ac{DC}, \ac{DDP}, and \ac{NMPC} on the Soft Cart-Pole, Soft Pendubot, and Soft Furuta Pendulum. These case studies confirmed the framework's capacity to handle non-minimum phase behaviors, complex out-of-plane deformations, and soft-link interactions. Crucially, the derivation of analytical gradients rendered the optimization of these high-order models computationally tractable; in their absence, such optimization problems would have been computationally prohibitive.

To ensure numerical stability and feasibility, we introduced \ac{Box-IDDP}, combining implicit time integration with input box constraints. 
Finally, we exploited the \ac{GVS} strain parameterization to implement a resolution-based warm-start strategy. By hierarchically increasing model fidelity, this approach enhanced convergence, which in turn led to a reduction in computational time.

Future work will focus on experimental validation of the benchmark systems, assessing real-time performance and robustness to model uncertainties.

\setcounter{equation}{0}
\renewcommand{\theequation}{\Alph{section}\arabic{equation}}

\appendix
\label{appendices:box_iddp_derivatives}

Let $\bm{\nabla}_{\bm{x}} \bm{f}\left(\bm{x}, \bm{u}\right) \in \mathbb{R}^{2n \times 2n}$ and $\bm{\nabla}_{\bm{u}} \bm{f}\left(\bm{x}, \bm{u}\right) \in \mathbb{R}^{2n \times m}$ be the analytical derivatives of the continuous-time dynamics, and let $\bm{g}\left(\bm{x}_k, \bm{x}_{k+1}, \bm{u}_k\right) = \bm{0}$ be the implicit discrete-time system. 
For the sake of readability, we refer to $\bm{x}_{k+1}$ as $\bm{x}'$, while $\bm{x}_k$ and $\bm{u}_k$ will be referred to as $\bm{x}$ and $\bm{u}$, respectively.
For instance, we denote $\bm{g}_{\bm{x}\bm{x}'} = \partial^{2}\bm{g} / \partial \bm{x}_k \bm{x}_{k+1}$.

The first- and second-order derivatives of the $Q$-function are listed below.
\begin{equation} \label{eq:Qx_Qu}
    \bm{Q}_{\bm{x}} = \bm{\ell}_{\bm{x}} + \mathbf{f}_{\bm{x}}^{\top} \bm{V}'_{\bm{x}'} , \quad \bm{Q}_{\bm{u}} = \bm{\ell}_{\bm{u}} + \mathbf{f}_{\bm{u}}^{\top} \bm{V}'_{\bm{x}'} \, , 
\end{equation}
\begin{equation} \label{eq:Qxx}
    \bm{Q}_{\bm{xx}} = \bm{\ell}_{\bm{xx}} + \mathbf{f}_{\bm{x}}^{\top} \bm{V}'_{\bm{x}'\bm{x}'} \mathbf{f}_{\bm{x}} + \bm{V}'_{\bm{x}'} \mathbf{f}_{\bm{xx}} \, , 
\end{equation}
\begin{equation} \label{eq:Quu}
    \bm{Q}_{\bm{uu}} = \bm{\ell}_{\bm{uu}} + \mathbf{f}_{\bm{u}}^{\top} \bm{V}'_{\bm{x}'\bm{x}'} \mathbf{f}_{\bm{u}} + \bm{V}'_{\bm{x}'} \mathbf{f}_{\bm{uu}} \, , 
\end{equation}
\begin{equation} \label{eq:Qux}
    \bm{Q}_{\bm{ux}} = \bm{\ell}_{\bm{ux}} + \mathbf{f}_{\bm{u}}^{\top} \bm{V}'_{\bm{x}'\bm{x}'} \mathbf{f}_{\bm{x}} + \bm{V}'_{\bm{x}'}  \mathbf{f}_{\bm{ux}} \, , 
\end{equation}
\begin{equation} \label{eq:Vx}
    \bm{V}_{\bm{x}} = \bm{Q}_{\bm{x}} - \bm{Q}^{\top}_{\bm{ux}} \bm{Q}^{-1}_{\bm{uu}} \bm{Q}_{\bm{u}} \, ,
\end{equation}
\begin{equation} \label{eq:Vxx}
    \bm{V}_{\bm{xx}} = \bm{Q}_{\bm{xx}} - \bm{Q}^{\top}_{\bm{ux}} \bm{Q}^{-1}_{\bm{uu}} \bm{Q}_{\bm{ux}} \, .
\end{equation}

In \eqref{eq:Qx_Qu}-\eqref{eq:Vxx}, the first-order derivatives of the implicit discrete-time dynamics $\mathbf{f}_{\bm{x}} \in \mathbb{R}^{2n \times 2n}$ and $\mathbf{f}_{\bm{u}} \in \mathbb{R}^{2n \times m}$ can be computed as follows, according to \cite{chatzinikolaidis2021trajectory}.
\begin{equation}
    \mathbf{f}_{\bm{x}} = \bm{g}_{\bm{x}'}^{\dagger} \bm{g}_{\bm{x}} \, , \quad \mathbf{f}_{\bm{u}} = \bm{g}_{\bm{x}'}^{\dagger} \bm{g}_{\bm{u}} \, ,
\end{equation}
where $\bm{g}_{\bm{x}'}^{\dagger} = - \bm{g}^{-1}_{\bm{x}'}$. Moreover, the second-order derivatives of the discrete-time dynamics $\mathbf{f}_{\bm{xx}} \in \mathbb{R}^{2n \times 2n \times 2n}$, $\mathbf{f}_{\bm{xu}} \in \mathbb{R}^{2n \times 2n \times m}$, and $\mathbf{f}_{\bm{uu}} \in \mathbb{R}^{2n \times m \times m}$ are shown below.
\begin{equation}
    \mathbf{f}_{\bm{xx}} = \bm{g}_{\bm{x}'}^{\dagger} \left(\mathbf{f}^{\top}_{\bm{x}} \bm{g}_{\bm{x'x'}}\mathbf{f}_{\bm{x}} + \mathbf{f}^{\top}_{\bm{x}} \bm{g}_{\bm{x'x}} + \bm{g}_{\bm{x'x}}^{\top} \mathbf{f}_{\bm{x}} + \bm{g}_{\bm{xx}}\right) ,
\end{equation}
\begin{equation}
    \mathbf{f}_{\bm{uu}} = \bm{g}_{\bm{x}'}^{\dagger} \left(\mathbf{f}^{\top}_{\bm{u}} \bm{g}_{\bm{x'x'}}\mathbf{f}_{\bm{u}} + \mathbf{f}^{\top}_{\bm{u}} \bm{g}_{\bm{x'u}} + \bm{g}_{\bm{x'u}}^{\top} \mathbf{f}_{\bm{u}} + \bm{g}_{\bm{uu}}\right) ,
\end{equation}
\begin{equation}
    \mathbf{f}_{\bm{xu}} = \bm{g}_{\bm{x}'}^{\dagger} \left(\mathbf{f}^{\top}_{\bm{x}} \bm{g}_{\bm{x'x'}}\mathbf{f}_{\bm{u}} + \mathbf{f}^{\top}_{\bm{x}} \bm{g}_{\bm{x'u}} + \bm{g}_{\bm{x'x}}^{\top} \mathbf{f}_{\bm{u}} + \bm{g}_{\bm{xu}}\right) .
\end{equation}

\bibliographystyle{IEEEtran}
\bibliography{references}

\end{document}


\title{Model-based Optimal Control for Rigid-Soft Underactuated Systems: \\ Supplementary Material}

\author{Author Names Omitted for Anonymous Review. Paper-ID 681}



\maketitle
\setcounter{equation}{0}
\renewcommand{\theequation}{\Alph{section}\arabic{equation}}

\appendix
\label{appendices:box_iddp_derivatives}

Let $\bm{\nabla}_{\bm{x}} \bm{f}\left(\bm{x}, \bm{u}\right) \in \mathbb{R}^{2n \times 2n}$ and $\bm{\nabla}_{\bm{u}} \bm{f}\left(\bm{x}, \bm{u}\right) \in \mathbb{R}^{2n \times m}$ be the analytical derivatives of the continuous-time dynamics, and let $\bm{g}\left(\bm{x}_k, \bm{x}_{k+1}, \bm{u}_k\right) = \bm{0}$ be the implicit discrete-time system. 
For the sake of readability, we refer to $\bm{x}_{k+1}$ as $\bm{x}'$, while $\bm{x}_k$ and $\bm{u}_k$ will be referred to as $\bm{x}$ and $\bm{u}$, respectively.
For instance, we denote $\bm{g}_{\bm{x}\bm{x}'} = \partial^{2}\bm{g} / \partial \bm{x}_k \bm{x}_{k+1}$.

The first- and second-order derivatives of the $Q$-function are listed below.
\begin{equation} \label{eq:Qx_Qu}
    \bm{Q}_{\bm{x}} = \bm{\ell}_{\bm{x}} + \mathbf{f}_{\bm{x}}^{\top} \bm{V}'_{\bm{x}'} , \quad \bm{Q}_{\bm{u}} = \bm{\ell}_{\bm{u}} + \mathbf{f}_{\bm{u}}^{\top} \bm{V}'_{\bm{x}'} \, , 
\end{equation}
\begin{equation} \label{eq:Qxx}
    \bm{Q}_{\bm{xx}} = \bm{\ell}_{\bm{xx}} + \mathbf{f}_{\bm{x}}^{\top} \bm{V}'_{\bm{x}'\bm{x}'} \mathbf{f}_{\bm{x}} + \bm{V}'_{\bm{x}'} \mathbf{f}_{\bm{xx}} \, , 
\end{equation}
\begin{equation} \label{eq:Quu}
    \bm{Q}_{\bm{uu}} = \bm{\ell}_{\bm{uu}} + \mathbf{f}_{\bm{u}}^{\top} \bm{V}'_{\bm{x}'\bm{x}'} \mathbf{f}_{\bm{u}} + \bm{V}'_{\bm{x}'} \mathbf{f}_{\bm{uu}} \, , 
\end{equation}
\begin{equation} \label{eq:Qux}
    \bm{Q}_{\bm{ux}} = \bm{\ell}_{\bm{ux}} + \mathbf{f}_{\bm{u}}^{\top} \bm{V}'_{\bm{x}'\bm{x}'} \mathbf{f}_{\bm{x}} + \bm{V}'_{\bm{x}'}  \mathbf{f}_{\bm{ux}} \, , 
\end{equation}
\begin{equation} \label{eq:Vx}
    \bm{V}_{\bm{x}} = \bm{Q}_{\bm{x}} - \bm{Q}^{\top}_{\bm{ux}} \bm{Q}^{-1}_{\bm{uu}} \bm{Q}_{\bm{u}} \, ,
\end{equation}
\begin{equation} \label{eq:Vxx}
    \bm{V}_{\bm{xx}} = \bm{Q}_{\bm{xx}} - \bm{Q}^{\top}_{\bm{ux}} \bm{Q}^{-1}_{\bm{uu}} \bm{Q}_{\bm{ux}} \, .
\end{equation}

In \eqref{eq:Qx_Qu}-\eqref{eq:Vxx}, the first-order derivatives of the implicit discrete-time dynamics $\mathbf{f}_{\bm{x}} \in \mathbb{R}^{2n \times 2n}$ and $\mathbf{f}_{\bm{u}} \in \mathbb{R}^{2n \times m}$ can be computed as follows, according to \cite{chatzinikolaidis2021trajectory}.
\begin{equation}
    \mathbf{f}_{\bm{x}} = \bm{g}_{\bm{x}'}^{\dagger} \bm{g}_{\bm{x}} \, , \quad \mathbf{f}_{\bm{u}} = \bm{g}_{\bm{x}'}^{\dagger} \bm{g}_{\bm{u}} \, ,
\end{equation}
where $\bm{g}_{\bm{x}'}^{\dagger} = - \bm{g}^{-1}_{\bm{x}'}$. Moreover, the second-order derivatives of the discrete-time dynamics $\mathbf{f}_{\bm{xx}} \in \mathbb{R}^{2n \times 2n \times 2n}$, $\mathbf{f}_{\bm{xu}} \in \mathbb{R}^{2n \times 2n \times m}$, and $\mathbf{f}_{\bm{uu}} \in \mathbb{R}^{2n \times m \times m}$ are shown below.
\begin{equation}
    \mathbf{f}_{\bm{xx}} = \bm{g}_{\bm{x}'}^{\dagger} \left(\mathbf{f}^{\top}_{\bm{x}} \bm{g}_{\bm{x'x'}}\mathbf{f}_{\bm{x}} + \mathbf{f}^{\top}_{\bm{x}} \bm{g}_{\bm{x'x}} + \bm{g}_{\bm{x'x}}^{\top} \mathbf{f}_{\bm{x}} + \bm{g}_{\bm{xx}}\right) ,
\end{equation}
\begin{equation}
    \mathbf{f}_{\bm{uu}} = \bm{g}_{\bm{x}'}^{\dagger} \left(\mathbf{f}^{\top}_{\bm{u}} \bm{g}_{\bm{x'x'}}\mathbf{f}_{\bm{u}} + \mathbf{f}^{\top}_{\bm{u}} \bm{g}_{\bm{x'u}} + \bm{g}_{\bm{x'u}}^{\top} \mathbf{f}_{\bm{u}} + \bm{g}_{\bm{uu}}\right) ,
\end{equation}
\begin{equation}
    \mathbf{f}_{\bm{xu}} = \bm{g}_{\bm{x}'}^{\dagger} \left(\mathbf{f}^{\top}_{\bm{x}} \bm{g}_{\bm{x'x'}}\mathbf{f}_{\bm{u}} + \mathbf{f}^{\top}_{\bm{x}} \bm{g}_{\bm{x'u}} + \bm{g}_{\bm{x'x}}^{\top} \mathbf{f}_{\bm{u}} + \bm{g}_{\bm{xu}}\right) .
\end{equation}





\bibliographystyle{plainnat}
\bibliography{references}